\documentclass{article}
\usepackage{graphicx} 
\usepackage{amsmath, amsthm, amssymb}
\usepackage{url}
\usepackage{geometry}
\title{Temporal Model On Quantum Logic}
\author{Francesco D'Agostino}
\date{\today}

\begin{document}

\maketitle

\begin{abstract}
This paper is the next part of the series of papers about the new quantum logic framework that I have developed. This paper deals with modeling temporal memory dynamics, combining concepts from temporal logic, memory decay models, and hierarchical contexts. We formalize the evolution of propositions over time using linear and branching temporal models, incorporating exponential decay (Ebbinghaus forgetting curve) and reactivation mechanisms via Bayesian updating. The hierarchical organization of memory is represented using directed acyclic graphs to model recall dependencies and interference.
\end{abstract}

\maketitle

\section{Temporal Logic Extensions}

Let $t \in \mathbb{R}$ represent a temporal parameter, and for any element $x \in \mathcal{S}$ at time $t$, we denote this as $x(t)$. The complete temporal state space is defined as $\mathcal{T}(x) = \{x(t) \mid t \in \mathbb{R}\}$, encompassing all possible states across the temporal domain. The evolution of any $x(t)$ occurs by a temporal transition function $f : \mathcal{T}(x) \times \mathbb{R} \to \mathcal{T}(x)$ such that $x(t + \Delta t) = f(x(t), \Delta t)$ for any temporal increment $\Delta t$.
An element $x$ evolves in linear time if and only if $t$ progresses along a single deterministic timeline, formally expressed as $t_1 < t_2 \implies x(t_1) \text{ precedes } x(t_2)$ for all $t_1, t_2 \in \mathbb{R}$. This establishes the temporal ordering of states. For elements evolving in linear time: $x(t)$ is unique for all $t \in \mathbb{R}$, which can be stated as $\forall t_1, t_2 \in \mathbb{R}, \quad t_1 \neq t_2 \implies x(t_1) \neq x(t_2)$. 
For this result to be established, the mapping $f: \mathbb{R} \rightarrow \mathcal{T}(x)$ defined by $f(t) = x(t)$ must be injective. So consider two distinct time points $t_1, t_2 \in \mathbb{R}$ such that $t_1 \neq t_2$. Without loss of generality, assume $t_1 < t_2$. Then, logically if $t_1 \neq t_2$, then either $t_1 < t_2$ or $t_2 < t_1$. Since we have  $t_1 < t_2$, this establishes a strict ordering between the two time points. According to what we said initially: if $t_1 < t_2$, then $x(t_1)$ precedes $x(t_2)$. The precedence relation in the temporal state space $\mathcal{T}(x)$ establishes that if $x(t_1)$ precedes $x(t_2)$, then $x(t_1) \neq x(t_2)$, as precedence is a strict partial order which is irreflexive. Therefore, for any $t_1, t_2 \in \mathbb{R}$ with $t_1 \neq t_2$, it follows that $x(t_1) \neq x(t_2)$, establishing that the mapping $t \mapsto x(t)$ is injective. This injectivity directly implies that $x(t)$ is unique for each $t \in \mathbb{R}$.
Now it is possible to introduce temporal operators that formalize reasoning across time. The operator $\Box x$ (Always) holds if and only if $x(t)$ is true for all $t \in \mathbb{R}$, formally expressed as $\Box x \iff \forall t \in \mathbb{R}, x(t) \text{ is true}$. The operator $\Diamond x$ (Eventually) holds if and only if $x(t)$ is true for some $t \in \mathbb{R}$, formally expressed as $\Diamond x \iff \exists t \in \mathbb{R}, x(t) \text{ is true}$. The operator $\bigcirc x$ (Next) holds if and only if $x(t + \Delta t)$ is true for a specific $\Delta t > 0$, formally expressed as $\bigcirc x \iff x(t + \Delta t) \text{ is true for a specific } \Delta t > 0$.
For any element $x$ in linear time, we can establish that $\Box x \implies \Diamond x$ and $\Diamond x \not\implies \Box x$. In order to prove the first implication, assume that $\Box x$ holds, meaning $\forall t \in \mathbb{R}, x(t) \text{ is true}$. From the axioms of first-order logic, if a property holds for all elements in a non-empty set, then there exists at least one element in that set for which the property holds. Since $\mathbb{R}$ is non-empty, the statement $\forall t \in \mathbb{R}, x(t) \text{ is true}$ implies $\exists t \in \mathbb{R}, x(t) \text{ is true}$, which by definition means $\Diamond x$ holds. 
For the second implication, let us consider a counterexample. Consider an element $x$ such that $x(t) = \begin{cases} \text{true}, & \text{if } t = t_0 \\ \text{false}, & \text{if } t \neq t_0 \end{cases}$ for some specific $t_0 \in \mathbb{R}$. For this element, $x(t_0)$ is true, which means $\exists t \in \mathbb{R}, x(t) \text{ is true}$, implying $\Diamond x$ holds. However, since $x(t)$ is false for all $t \neq t_0$, the statement $\forall t \in \mathbb{R}, x(t) \text{ is true}$ is false, meaning $\Box x$ does not hold. This provides a counterexample establishing that $\Diamond x \not\implies \Box x$.
It is also possible to prove that for all elements $x \in \mathcal{S}$ evolving in linear time, the implication $\bigcirc (\Box x) \implies \Box (\bigcirc x)$ holds. The left side of the implication, $\bigcirc (\Box x)$, means that $\Box x$ holds at the next time step, which can be expressed as $\bigcirc (\Box x) \iff (\Box x)(t + \Delta t) \text{ is true for a specific } \Delta t > 0$. Expanding this using the definition of $\Box$, we get $\bigcirc (\Box x) \iff \forall s \in \mathbb{R}, x(s) \text{ is true at time } t + \Delta t$. Since $x$ evolves in linear time, the temporal parameter $s$ must be interpreted relative to the current time $t + \Delta t$, giving us $\bigcirc (\Box x) \iff \forall s \in \mathbb{R}, x(t + \Delta t + s) \text{ is true}$. 
The right side of the implication, $\Box (\bigcirc x)$, means that $\bigcirc x$ holds at all times, which can be expressed as $\Box (\bigcirc x) \iff \forall u \in \mathbb{R}, (\bigcirc x)(u) \text{ is true}$. Expanding this using the definition of $\bigcirc$, we get $\Box (\bigcirc x) \iff \forall u \in \mathbb{R}, x(u + \Delta t) \text{ is true for a specific } \Delta t > 0$, which can be rewritten as $\Box (\bigcirc x) \iff \forall v \in \mathbb{R} \text{ where } v = u + \Delta t, x(v) \text{ is true}$. Since $u$ ranges over all of $\mathbb{R}$ and $\Delta t$ is a fixed positive constant, $v = u + \Delta t$ also ranges over all of $\mathbb{R}$, giving us $\Box (\bigcirc x) \iff \forall v \in \mathbb{R}, x(v) \text{ is true}$. 
To establish the implication, assume that $\bigcirc (\Box x)$ holds, meaning $\forall s \in \mathbb{R}, x(t + \Delta t + s) \text{ is true}$. For any arbitrary $v \in \mathbb{R}$, we can express $v$ as $v = t + \Delta t + s$ for some $s \in \mathbb{R}$ by setting $s = v - (t + \Delta t)$. Since $\bigcirc (\Box x)$ holds, $x(t + \Delta t + s)$ is true for all $s \in \mathbb{R}$, including $s = v - (t + \Delta t)$. Therefore, $x(v)$ is true for all $v \in \mathbb{R}$, which by definition means that $\Box (\bigcirc x)$ holds.
Extending beyond linear time, let us introduce a branching model that incorporates multiple possible futures. This branching model introduces $\mathcal{B}(t)$, a branching set of states defined as $\mathcal{B}(t) = \{x_b(t) \mid b \in \mathcal{B}, \, \mathcal{B} \text{ is the set of branches}\}$. In this model, $x(t)$ at time $t$ branches into multiple states, expressed as $x(t) = \bigcup_{b \in \mathcal{B}} x_b(t)$, where $\mathcal{B}$ is the set of all branches. A branch $b \in \mathcal{B}$ is realized if and only if there exists at least one time point $t \in \mathbb{R}$ such that the measurement $M$ applied to $x_b(t)$ in branch $b$ at time $t$ yields a result different from the bottom element $\bot$, formally expressed as $\exists t \in \mathbb{R}, \, M(x_b(t)) \neq \bot$. 
The measurement operation $M$ is a function that maps an element in a specific branch at a specific time to either a resolved state or the bottom element $\bot$, defined as $M: \mathcal{T}(x_b) \rightarrow \mathcal{S} \cup \{\bot\}$, where $\mathcal{T}(x_b)$ is the temporal state space of element $x$ in branch $b$, and $\mathcal{S}$ is the set of all elements. The bottom element $\bot$ represents an unresolved or indeterminate state. If $M(x_b(t)) = \bot$, this indicates that the measurement at branch $b$ at time $t$ does not yield a definite result, meaning the state remains indeterminate in that branch at that time. Conversely, if for all time points $t \in \mathbb{R}$, the measurement in branch $b$ yields the bottom element, i.e., $\forall t \in \mathbb{R}, \, M(x_b(t)) = \bot$, then the branch $b$ remains unrealized.
If an element $x(t)$ exists across multiple branches at time $t$, then it exists in a state of superposition that can be formally expressed as $x(t) = \bigcup_{b \in \mathcal{B}} \{x_b(t)\}$. The inclusion of braces around $x_b(t)$ indicates that each branch-specific state is being treated as a singleton set. The union of these singleton sets yields a set containing all the branch-specific states as distinct elements, rather than potentially merging them as would be the case with $\bigcup_{b \in \mathcal{B}} x_b(t)$ if $x_b(t)$ were itself a set. This is crucial for the concept of superposition, analogous to quantum mechanics where a system can exist in multiple states simultaneously until measured. In this context, the element $x$ at time $t$ exists simultaneously in all its branch-specific states $x_b(t)$ until a measurement or observation collapses it into a specific branch. Each branch-specific state $x_b(t)$ represents a distinct possibility at time $t$, and the element exists in all these possibilities simultaneously until a measurement causes it to collapse into a specific branch.

\section{Realization and Decay Dynamics}

From this framework that has being developed, it is necessary to expand on the concept of realization and decay dynamics, in order to understand how states evolve over time. For any element $P \in \mathcal{S}$, two distinct temporal states can be identified: the realized state and the decayed memory state.
The realized state of $P$ occurs during the interval $[t_r, t_f)$ and is denoted by $P(!t)$ for $t \in [t_r, t_f)$, where $t_r$ represents the time of realization and $t_f$ marks the end of this realization. Following this period, at times $t \geq t_f$, the relationship weakens and transitions to a decayed memory state denoted by $P(?t)$.
The temporal evolution of $P$ can be formally expressed through a piecewise function:
\begin{equation}
P(t) = 
\begin{cases} 
P(!t), & t \in [t_r, t_f), \\ 
P(?t), & t \geq t_f. 
\end{cases} 
\end{equation}
This formulation follows directly what we established earlier. According to the temporal state parameter definition, $P(t)$ denotes the state at time $t$, acknowledging that elements exist within a temporal framework and can assume different states at different points in time. The transition between states occurs precisely at $t_f$, with no gaps in the temporal evolution.
The strength of the decayed memory state $P(?t)$ is governed by a decay function $d(t): [t_f, \infty) \to [0, 1]$, where $d(t_f) = 1$ and $\lim_{t \to \infty} d(t) = 0$. This gives us $\text{Strength}(P(?t)) = d(t)$ for $t \geq t_f$. 
A reactivation function $r(C_m, t)$ determines whether $P(?t)$ can transition back to a realized state under a triggering context $C_m$. Formally:
\begin{equation}
r(C_m, t) = 
\begin{cases} 
1, & \text{if } C_m \neq \varnothing, \\ 
0, & \text{if } C_m = \varnothing. 
\end{cases} 
\end{equation}
When $C_m \neq \varnothing$, the decayed state transitions back to a realized state: $P(?t) \xrightarrow{C_m} P(!t')$ for some $t' > t_f$. On the other hand, if $C_m = \varnothing$, $P(?t)$ remains in the decayed state and approaches $\bot$ as $t \to \infty$.
The dynamics can be further refined by expressing $P(t)$ as:
\begin{equation}
P(t) = 
\begin{cases} 
P(!t), & t \in [t_r, t_f), \\ 
P(?t) = d(t)P_m, & t \geq t_f. 
\end{cases} 
\end{equation}
This formulation incorporates the decay function directly into the state representation. For temporal operators, $\Diamond P$ holds if there exists some time $t \geq t_f$ where $d(t) > 0$. Correspondingly, if $d(t) = 0$ as $t \to \infty$, then $\Box P(?t) = 0$ as $t \to \infty$, indicating that the always operator cannot hold for a completely decayed state.
A specific implementation of the decay function based on Ebbinghaus' forgetting curve is given by $d_E(t) = e^{-\lambda(t - t_f)}$, where $\lambda > 0$ is the decay constant determining the rate of memory fading, $t_f$ is the time at which realization ends, and $t \geq t_f$ is the time of observation. For a state governed by this function, the strength is $\text{Strength}(P(?t)) = e^{-\lambda(t - t_f)}$ for $t \geq t_f$, with $\lim_{t \to \infty} \text{Strength}(P(?t)) = 0$.
When a triggering context $C_m \neq \varnothing$ causes reactivation at time $t' > t_f$, the strength is restored to a realized state: $P(?t) \xrightarrow{C_m} P(!t')$ with $\text{Strength}(P(!t')) = 1$. If the state decays following $d_E(t)$, repeated reactivation at intervals $t_k$ resets the decay function, formally expressed as $t_k > t_{k-1} \implies d_E(t_k) = 1$ for all $k \in \mathbb{N}$. Without reactivation ($C_m = \varnothing$), the state follows the exponential decay pattern without interruption: $\text{Strength}(P(?t)) = e^{-\lambda(t - t_f)}$ as $t \to \infty$.

\section{Memory Interference and Hierarchical Contextual Dynamics}

This study can be extended to incorporate memory interference and hierarchical contextual dynamics, which provides a more comprehensive understanding of how propositions interact within complex memory structures. Let us begin by considering the set of all propositions $\mathcal{S}$ and defining a memory chain as a subset $\{P_1, P_2, \dots, P_n\} \subset \mathcal{S}$ where there exists a relation $R : \mathcal{S} \times \mathcal{S} \to [0, 1]$ such that $R(P_i, P_j) > 0$ for all $i, j \in \{1, 2, \dots, n\}$. This relation quantifies the strength of association between any two propositions in the memory chain.
To capture this nested contextual information nature, we need to introduce hierarchical context structures. A hierarchical context $\mathcal{C}$ is defined as a collection $\{C_1, C_2, \dots\}$ where each $C_i \subseteq C_{i+1}$ represents a subcontext contained within a broader context. This nesting allows the modeling of how memory operates at different levels of abstraction. A proposition $P$ belongs to a specific context $C$ if there exists some $i$ such that $P \in C_i$ and $C_i \in \mathcal{C}$. This formal definition captures the intuitive notion that propositions exist within specific contextual frameworks.
The contextual degree of relation between two propositions $P_i$ and $P_j$ within a shared context $C_k$ is given by $R_C(P_i, P_j)$, which equals $R(P_i, P_j)$ if both propositions belong to $C_k$, and 0 otherwise. This definition allows the quantification of how strongly related two propositions are within a specific context, which is essential for modeling context-dependent memory dynamics.
When propositions in $\mathcal{C}$ form a hierarchical memory chain, we have $R_C(P_i, P_j) > 0$ for all $P_i, P_j \in C_k$ and $C_k \subseteq \mathcal{C}$. This means that within any given context in the hierarchy, all propositions have some degree of relation to each other. Furthermore, if a proposition $P_i$ belongs to a context $C_k$ which is a subset of the broader context structure $\mathcal{C}$, then the recall of $P_i$ propagates to another proposition $P_j$ in broader contexts $C_l$ (where $l > k$) whenever $R_C(P_i, P_j) > \tau$, where $\tau$ is a threshold value. This propagation mechanism models how activating one memory can trigger related memories in broader contextual conditions.
From this hierarchical structure, entanglement between propositions can also occur. Two propositions $P_i$ and $P_j$ within a shared context $C_k$ are considered entangled if $R_C(P_i, P_j) > \tau_e$, where $\tau_e$ is the entanglement threshold. This entanglement is: $P_i \triangleleft P_j$. A key property of this entanglement is that it persists across the hierarchy: if $P_i \triangleleft P_j$ in a lower context $C_k$, then $P_i \triangleleft P_j$ in all higher contexts $C_l$ where $l > k$. This persistence captures how strongly associated memories remain connected regardless of the breadth of context being considered.
To visualize and analyze these hierarchical structures more effectively, we can represent them as directed acyclic graphs. The hierarchical structure $\mathcal{C}$ is represented as a graph $G(\mathcal{C}) = (\mathcal{V}, \mathcal{E})$, where the vertices $\mathcal{V} = \{C_k \mid C_k \in \mathcal{C}\}$ correspond to contexts, and the edges $\mathcal{E} = \{(C_k, C_l) \mid C_k \subseteq C_l\}$ represent the containment relationships between contexts. Within this graph representation, the recall of a proposition $P_i$ propagates through the graph along paths $\pi$, with the probability of recalling another proposition $P_j$ given by $P(\text{recall } P_j) = \sum_{\pi \in G(\mathcal{C})} \prod_{(C_k, C_l) \in \pi} R_C(P_i, P_j)$. This formula accounts for all possible paths through which activation can spread in the contextual hierarchy. A direct consequence of this propagation mechanism is that if $P_i \in C_k$ and $P_j \in C_l$ with $C_k \subseteq C_l$, then the probability of recalling $P_j$ is positive, i.e., $P(\text{recall } P_j) > 0$.
The temporal aspects of memory recall can be captured through the concept of recall latency. The recall latency $T_R(P_i)$ represents the time required for a proposition $P_i$ in a memory state, denoted as $P_i(?t)$, to transition to an actuated state of anamnesis, denoted as $P_i(!t)$. This latency depends on two key factors: the degree of relation $R_C(P_i, P_j)$ between $P_i$ and an already recalled proposition $P_j$, and environmental factors $\mathcal{E}$ that influence the recall process. When the relation strength between propositions varies, so does the recall latency. Specifically, if $R_C(P_i, P_j) > R_C(P_i, P_k)$, then $T_R(P_i \mid P_j) < T_R(P_i \mid P_k)$, meaning that propositions more strongly related to already recalled propositions are recalled more quickly.
The environmental facilitation factor $\mathcal{E}$ modifies recall latency according to the formula $T_R(P_i) = \frac{T_B(P_i)}{\mathcal{E}}$, where $T_B(P_i)$ is the base latency and $\mathcal{E} > 0$ can either amplify or diminish the recall time. These environmental circumstances $\mathcal{E}$ are defined as a function of external stimuli or context $C_{\text{external}}$ and internal cognitive states $C_{\text{internal}}$ such as focus or emotional relevance: $\mathcal{E} = f(C_{\text{external}}, C_{\text{internal}})$. When $\mathcal{E} > 1$, recall latency is reduced below the base latency, i.e., $T_R(P_i) < T_B(P_i)$. Conversely, when $\mathcal{E} < 1$, recall latency increases above the base latency, i.e., $T_R(P_i) > T_B(P_i)$.
Exceptional recall events can occur when a proposition $P_k$ with a weak relation $R_C(P_i, P_k)$ transitions rapidly due to extremely high environmental facilitation. In such cases, as $\mathcal{E}$ approaches infinity, the recall latency $T_R(P_k)$ approaches the recall latency of strongly related propositions: $T_R(P_k) \to T_R(P_j)$ as $\mathcal{E} \to \infty$. In the limit, as $\mathcal{E} \to \infty$, recall latency becomes independent of the relation strength $R_C(P_i, P_k)$ and approaches zero for all propositions: $\lim_{\mathcal{E} \to \infty} T_R(P_k) = 0$ for all $P_k$. For propositions with low relation strength $R_C(P_i, P_k)$, the recall latency follows a nonlinear function $T_R(P_k) = g(R_C(P_i, P_k), \mathcal{E})$, where $g$ is nonlinear as $R_C(P_i, P_k)$ approaches zero.
Two important special cases arise from these dynamics. First, when the relation strength is very small ($R_C(P_i, P_k) \ll 1$) but environmental facilitation is very high ($\mathcal{E} \gg 1$), recall latency approaches zero: $R_C(P_i, P_k) \ll 1 \land \mathcal{E} \gg 1 \implies T_R(P_k) \to 0$. Second, when relation strength approaches zero ($R_C(P_i, P_k) \to 0$) and environmental facilitation is approximately normal ($\mathcal{E} \approx 1$), recall latency diverges to infinity: $R_C(P_i, P_k) \to 0 \land \mathcal{E} \approx 1 \implies T_R(P_k) \to \infty$.
The transition from a memory state $P(?t)$ to an actuated state $P(!t)$ occurs over the recall latency $T_R(P_i)$, which we can express as $P(?t) \xrightarrow{T_R(P_i)} P(!t)$. The recall latency is determined by the formula:
\begin{equation}
T_R(P_i) = \frac{1}{R_C(P_i, P_j) \cdot \mathcal{E}},  
\end{equation}
where $R_C(P_i, P_j)$ is the degree of relation to an already recalled proposition $P_j$, and $\mathcal{E}$ is the environmental facilitation factor. The probability of transitioning from $P(?t)$ to $P(!t)$ at time $t$ follows an exponential distribution: 
\begin{equation}
P(\text{transition at } t) = 1 - e^{-\frac{t}{T_R(P_i)}}.
\end{equation}
This formula captures the increasing likelihood of recall as time passes.
The relationship between relation strength and transition speed is such that if $R_C(P_i, P_j) > R_C(P_i, P_k)$, then $T_R(P_i \mid P_j) < T_R(P_i \mid P_k)$. Similarly, for a fixed relation strength $R_C(P_i, P_j)$, higher environmental facilitation reduces transition latency: if $\mathcal{E}_1 > \mathcal{E}_2$, then $T_R(P_i \mid \mathcal{E}_1) < T_R(P_i \mid \mathcal{E}_2)$.
Propositions can be organized into temporal hierarchies based on their recall latencies. A proposition $P_i$ belongs to a temporal hierarchy $\mathcal{H}(P)$ if its recall latency is below a given threshold $\epsilon$: $\mathcal{H}(P) = \{P_i \mid T_R(P_i) \leq \epsilon\}$. Within such a hierarchy, transitions occur in order of increasing recall latency: if $T_R(P_1) < T_R(P_2) < \cdots < T_R(P_n)$, then the recall sequence follows $P_1 \to P_2 \to \cdots \to P_n$. However, this orderly progression can be interrupted by environmental facilitation. A proposition $P_k$ that does not belong to the hierarchy $\mathcal{H}(P)$ may transition earlier than a proposition $P_j$ within the hierarchy if the environmental facilitation for $P_k$ is much greater than for $P_j$: $\mathcal{E}_k \gg \mathcal{E}_j \implies T_R(P_k) < T_R(P_j)$.
The state of a proposition $P_i$ at time $t$, denoted $P_i(t)$, evolves according to a unified transition dynamic: $P_i(t) = P(?t)$ if $t < T_R(P_i)$, and $P_i(t) = P(!t)$ if $t \geq T_R(P_i)$. When a temporal hierarchy $\mathcal{H}(P)$ contains multiple propositions, all propositions with recall latencies greater than the current time remain in the memory state: $P_i(t) = P(?t)$ for all $i$ with $T_R(P_i) > t$.
The organizational efficiency of memory chains can be quantified using entropy. The entropy $H(\mathcal{C})$ of a memory chain $\mathcal{C} = \{P_1, P_2, \dots, P_n\}$ is defined as $H(\mathcal{C}) = -\sum_{i=1}^n p(P_i) \log p(P_i)$, where $p(P_i)$ is the probability of recalling $P_i$ within $\mathcal{C}$. When the average relation strength $R_C(P_i, P_j)$ in $\mathcal{C}$ is high, the entropy $H(\mathcal{C})$ is low, indicating a well-organized memory structure. Conversely, when $R_C(P_i, P_j)$ is low, $H(\mathcal{C})$ increases, reflecting a more disorganized memory structure. Higher entropy in $\mathcal{C}$ corresponds to longer average recall latency $T_R(\mathcal{C})$, with the relationship being proportional: $H(\mathcal{C}) \propto \frac{1}{n} \sum_{i=1}^n T_R(P_i)$.
The simultaneous influence of propositions on each other is another important aspect of memory dynamics. For a set of propositions $\mathcal{S} = \{P_1, P_2, \dots, P_n\}$, the simultaneous influence $\mathcal{I}(P_i, P_j)$ of $P_i$ on $P_j$ is defined as $\mathcal{I}(P_i, P_j) = R_C(P_i, P_j) \cdot \frac{1}{T_R(P_i)}$. When propositions $P_i$ and $P_j$ are entangled, their mutual influence is symmetric: $\mathcal{I}(P_i, P_j) = \mathcal{I}(P_j, P_i)$. The recall latency $T_R(P_i)$ is adjusted by the cumulative influence of all related propositions $P_j$ according to the formula:
\begin{equation}
T_R(P_i) = \frac{1}{\sum_{j \neq i} \mathcal{I}(P_j, P_i)}
\end{equation}
Dynamic feedback mechanisms play a crucial role in memory recall. The feedback influence $F(P_i, P_j)$ from $P_i$ to $P_j$ modifies the recall latency $T_R(P_j)$ when $P_i$ is recalled. This influence is modeled as $F(P_i, P_j) = \alpha \cdot R_C(P_i, P_j)$, where $\alpha$ is a feedback coefficient. When a proposition $P_i$ is recalled at time $t_r$, it can reset the Ebbinghaus decay curve of related propositions. The Ebbinghaus decay of $P_j$ resets to $d_E(P_j, t) = e^{-\lambda (t - t_r)}$, where $\lambda$ is the decay constant and $t - t_r$ is the time elapsed since the recall of $P_i$. The adjusted recall latency $T_R(P_j)$ after $P_i$ is recalled satisfies: 
\begin{equation}
T_R(P_j) = \frac{T_B(P_j)}{1 + F(P_i, P_j)},
\end{equation}
where $T_B(P_j)$ is the base latency. Repeated recall of $P_i$ creates a feedback loop for $P_j$, such that $T_R(P_j) \to 0$ as $F(P_i, P_j) \to \infty$.
Bayesian inference provides a powerful framework for modeling memory dynamics. The conditional probability $P(P_j \mid P_i)$ represents the likelihood of recalling $P_j$ given that $P_i$ has been recalled. The Bayesian update rule for this conditional probability is:
\begin{equation}
P(P_j \mid P_i) = \frac{P(P_i \mid P_j) P(P_j)}{P(P_i)}, 
\end{equation}
where $P(P_i \mid P_j)$ is the likelihood of recalling $P_i$ if $P_j$ is present, $P(P_j)$ is the prior probability of recalling $P_j$, and $P(P_i)$ is the evidence or marginal probability of $P_i$. The decay function for $P_j$ after $P_i$ is recalled, modified by Bayesian inference, is $d_B(P_j, t) = P(P_j \mid P_i) \cdot e^{-\lambda (t - t_r)}$, where $t_r$ is the recall time of $P_i$. The recall latency $T_R(P_j)$, after applying Bayesian inference with feedback from $P_i$, is:
\begin{equation}
T_R(P_j) = \frac{T_B(P_j)}{P(P_j \mid P_i)}.
\end{equation}
The posterior update for $P(P_j \mid P_i)$ after feedback is given by:
\begin{equation}
P'(P_j \mid P_i) = \frac{P(P_i \mid P_j) P(P_j)}{\int_{P_k \in \mathcal{C}} P(P_i \mid P_k) P(P_k) dP_k}. 
\end{equation}
If $P_i$ is recalled repeatedly, the updated probability $P'(P_j \mid P_i)$ increases with each iteration: $P'(P_j \mid P_i) \geq P(P_j \mid P_i)$. As feedback iterations $n$ approach infinity, the recall probability converges to 1 for strongly linked propositions: $\lim_{n \to \infty} P^{(n)}(P_j \mid P_i) = 1$.
The Bayesian feedback mechanism modifies the Ebbinghaus decay curve, increasing the persistence of $P_j$ in memory: $d_B(P_j, t) \geq d_E(P_j, t)$ for all $t \geq t_r$. As a result, the modified recall latency $T_R(P_j)$ under Bayesian feedback is always shorter than the baseline: $T_R(P_j) \leq T_B(P_j)$.
Causality and subjective memory organization are also important aspects of memory dynamics. The influence of a recalled proposition $P_i(!t)$ on a proposition $P_j(?t)$ in a memory state depends on whether they belong to the same memory chain. The chain influence $\mathcal{I}_c(P_i \to P_j)$ equals $R_C(P_i, P_j)$ if $P_i$ and $P_j$ are in the same chain, and $\epsilon R_U(P_i, P_j)$ if they belong to different chains, where $R_U(P_i, P_j)$ represents a universal relationship and $\epsilon \ll 1$ quantifies the subtlety of cross-chain links. If $P_i(!t)$ and $P_j(?t)$ belong to the same memory chain, the influence is causal: 
\begin{equation}
\mathcal{I}_c(P_i \to P_j) > 0 \implies P_j(?t) \xrightarrow{T_R(P_j)} P_j(!t).
\end{equation}
If they belong to different chains, the influence is non-causal: $\mathcal{I}_c(P_i \to P_j) = \epsilon R_U(P_i, P_j)$ with $\epsilon \ll 1$. The recall of $P_j(?t)$ depends on the subjective organization of memory chains: $P_j(?t) \xrightarrow{T_R(P_j)} P_j(!t)$ if $\mathcal{I}_c(P_i \to P_j) \geq \tau$.
It is important to note that this framework models the generation of propositions $P$, which precedes classical logic. Memory recall does not create truth but observes and organizes pre-logical structures: $M(P(?t)) \neq \bot$ implies observation, not creation. The proposed framework unifies memory dynamics and proposition generation, operating before classical logic applies: $P(?t) \to P(!t)$ implies that logical evaluation follows.
Hierarchical memory structures can introduce biases in recall. A memory recall is said to be imprecise if it differs from the original proposition. Imprecision $\delta(P_i)$ is defined as $\delta(P_i) = 1 - R_C(P_i, P_i')$, where $P_i'$ is the retrieved approximation of $P_i$ and $R_C$ quantifies their similarity. Hierarchical recall introduces exponential decay in $R_C(P_i, P_i')$ over time: $R_C(P_i, P_i') = e^{-\lambda t}$, where $\lambda > 0$ is the decay rate. As $t \to \infty$, recall strength diminishes to a state of "numbness," where $R_C(P_i, P_i') \to 0$ but $P_i$ remains in superposition: $R_C(P_i, P_i') = 0 \implies P_i(?t) = \{P_m, \bot\}$. Hierarchical memory structures prioritize propositions with higher initial $R_C$ values, introducing bias in recall: $P_i(?t) \to P_i(!t)$ if $R_C(P_i, P_j) \geq \tau$.
Recursive models of recall and indirect updates provide a more comprehensive understanding of memory dynamics. The influence network of a proposition $P_i$ in a set $\mathcal{S} = \{P_1, P_2, \dots, P_n\}$ is represented as a directed graph $G(P_i) = (\mathcal{V}, \mathcal{E})$, where $\mathcal{V} = \{P_j \in \mathcal{S} \mid R_C(P_i, P_j) > 0\}$ and $\mathcal{E} = \{(P_j, P_k) \mid R_C(P_j, P_k) > 0\}$. The recursive influence of $P_i$ on $P_j$, denoted $\mathcal{R}(P_i \to P_j)$, is the cumulative influence through all direct and indirect paths in $G(P_i)$: $\mathcal{R}(P_i \to P_j) = \sum_{\pi \in \Pi(P_i \to P_j)} \prod_{(P_k, P_l) \in \pi} R_C(P_k, P_l)$, where $\Pi(P_i \to P_j)$ is the set of all paths from $P_i$ to $P_j$ in $G(P_i)$. The total influence of $P_i(!t)$ on $P_j(?t)$ is $\mathcal{I}_T(P_i \to P_j) = R_C(P_i, P_j) + \mathcal{R}(P_i \to P_j)$, where $R_C(P_i, P_j)$ represents direct influence and $\mathcal{R}(P_i \to P_j)$ accounts for indirect influence. For paths $\pi \in \Pi(P_i \to P_j)$, the influence along longer paths decays exponentially: $\prod_{(P_k, P_l) \in \pi} R_C(P_k, P_l) \propto e^{-\lambda |\pi|}$, where $|\pi|$ is the length of the path and $\lambda > 0$ is the decay constant.
The updated recall probability of $P_j$ after recalling $P_i$ is denoted $P'(P_j \mid P_i)$, incorporating both direct and recursive influences: 
\begin{equation}
P'(P_j \mid P_i) = \frac{\mathcal{I}_T(P_i \to P_j)}{\sum_{P_k \in \mathcal{V}} \mathcal{I}_T(P_i \to P_k)}. 
\end{equation}
This recall probability is proportional to the combined direct and indirect influences: $P'(P_j \mid P_i) \propto R_C(P_i, P_j) + \mathcal{R}(P_i \to P_j)$.
Recursive feedback loops occur when recalling $P_i(!t)$ indirectly reinforces its own influence through other propositions: $F_R(P_i) = \sum_{P_j \in \mathcal{V}} \mathcal{R}(P_j \to P_i)$. The feedback-adjusted recall probability of 
$P_i$ is:
\begin{equation}
 P''(P_i) = \frac{P'(P_i) + F_R(P_i)}{1 + \sum_{P_j \in \mathcal{V}} F_R(P_j)}.
\end{equation}
As feedback iterations increase, recall probabilities converge: $\lim_{n \to \infty} P^{(n)}(P_j \mid P_i) = 1$ for strongly linked propositions $P_j$.
Memory resilience and decay resistance are important properties of memory systems. The resilience of a proposition $P_i$, denoted $\mathcal{R}_i$, is defined as the cumulative effect of recall events $t_k$ on its decay rate $\lambda_i$: $\mathcal{R}_i = \sum_{k=1}^n e^{-\alpha (t_k - t_{k-1})}$, where $\alpha > 0$ is the resilience decay constant, and $t_k$ are the times of successive recalls. The adjusted decay rate $\lambda_i'$ for $P_i$ is inversely proportional to its resilience: 
\begin{equation}
\lambda_i' = \frac{\lambda_i}{1 + \mathcal{R}_i}. 
\end{equation}
If recall events occur with increasing frequency such that $t_k - t_{k-1} \to 0$ as $k \to \infty$, then resilience approaches infinity and the adjusted decay rate approaches zero: $\lim_{k \to \infty} \mathcal{R}_i \to \infty \implies \lim_{k \to \infty} \lambda_i' \to 0$. For resilience values above a threshold $\tau$, decay becomes negligible: $\mathcal{R}_i \geq \tau \implies \lambda_i' \approx 0$.
Finally, entropy plays a crucial role in recall efficiency. The entropy $H(\mathcal{C})$ of a memory chain $\mathcal{C} = \{P_1, P_2, \dots, P_n\}$ quantifies its organizational efficiency: $H(\mathcal{C}) = -\sum_{i=1}^n p(P_i) \log p(P_i)$, where $p(P_i)$ is the recall probability of $P_i$. The recall efficiency $E(\mathcal{C})$ of $\mathcal{C}$ is inversely related to its entropy: $E(\mathcal{C}) = \frac{1}{1 + H(\mathcal{C})}$. The optimal recall distribution minimizes $H(\mathcal{C})$: $p(P_i) = \frac{1}{Z} e^{-\beta R_C(P_i, P_j)}$, where $Z$ is the normalization factor, $\beta > 0$ is the entropy scaling parameter, and $R_C(P_i, P_j)$ is the relation strength. For low-entropy chains ($H(\mathcal{C}) \ll 1$), recall latency $T_R(P_i)$ is minimized: $H(\mathcal{C}) \ll 1 \implies T_R(P_i) \to \min$. Conversely, if entropy approaches infinity ($H(\mathcal{C}) \to \infty$), recall precision decreases: $H(\mathcal{C}) \to \infty \implies \delta(P_i) \to 1$.

\section{Conclusion}
This paper presents a study for temporal memory dynamics, integrating temporal logic, Ebbinghaus decay, and Bayesian reactivation within a hierarchical context. It is aimed at demonstrating the progression of propositions between realized and decayed states and introduced mechanisms for feedback-driven memory reinforcement. Hierarchical and recursive models provided insight into recall dependencies and entropy-based memory organization.

\end{document}